# Advanced Mathematics Learning Behavior Prediction and Academic Early Warning Model Based on Multimodal Data Analysis

Liu Qiong[1*] and Li Zhengbo[1]

[1] Moutai Institute , Renhuai, 564507, Guizhou, China

**Abstract.** Early detection of at-risk students and timely academic intervention are critical challenges in advanced mathematics education, were complex conceptual hierarchies and nonlinear learning trajectories often hinder student success. This study aims to address these issues by developing a dynamic framework for predicting learning behaviors and issuing academic early warnings using multimodal data analytics. The proposed approach constructs a hierarchical knowledge graph ontology, incorporates adaptive edge weighting based on problem difficulty and student performance, and integrates heterogeneous graph attention with temporal sequence modeling to capture students' evolving knowledge states. Empirical evaluation on semester-long multimodal datasets demonstrates robust identification of high-risk students, accurate tracing of error propagation, and significant improvements in knowledge mastery and risk reduction following targeted interventions. These findings indicate that integrating knowledge graph analytics with multimodal temporal modeling enables more effective and personalized learning support in advanced mathematics.

## 1 INTRODUCTION

The rapid evolution of educational data mining has transformed the analytical landscape of higher mathematics learning, enabling the extraction of complex behavioral patterns from diverse digital footprints [1]. As a foundational discipline in STEM education, advanced mathematics presents unique modeling challenges due to its abstract conceptual structure and the nonlinear progression of student understanding [2]. Although prior research has leveraged online activity logs and assessment data to investigate students' cognitive trajectories and knowledge acquisition, most studies are constrained by their reliance on unimodal or static data sources [3].

Because of these limitations, it is important to employ methods that can integrate multimodal educational data to provide a more comprehensive view of the learning process. Multimodal data analysis uses heterogeneous sources, such as text responses, clickstream interactions, semantic knowledge graphs, text responses, and time-assessed logs, to better understand the evolution of students' knowledge states [4]. Modeling complex semantic relationships between mathematical concepts is now possible thanks to advances in graph-based representation learning techniques. This helps in distinguishing conceptual interdependencies and distinctions to an unprecedented degree [5]. Temporal analysis also allows teachers to monitor changes in student mastery, which can help in the early detection of problems and the development of individualized interventions for students [6]. However, existing methods often fail to model failure and knowledge dissemination because conceptual networks are constantly changing [7]. Therefore, there is a need for dynamic schema-driven frameworks that can provide actionable support and better predict academic risk [8].

In this work, a multimodal, graph-based framework is proposed for predicting advanced mathematics learning behaviors and issuing academic early warnings. The presented technique addresses the main issues of previous research by constructing a high-dimensional knowledge graph ontology, introducing an edge-weighting mechanism based on problem difficulty, and integrating a sequence model based on graph neural networks. The effectiveness and accuracy of the models are demonstrated through case studies and empirical validation. Furthermore, the approach shows potential for enhanced individualized support and early risk mitigation in mathematics education.

## 2 Knowledge Graph Construction

### 2.1 Concept Ontology Engineering

Learning analytics for advanced mathematics requires a solid foundation that can identify hierarchical structures and interrelationships among mathematical concepts. Fine-grained knowledge tracking and scalable analysis can be achieved by systematically modeling the preconditioning and postconditioning relationships between core topics (e.g., sets, functions, derivatives, integrals, and linear algebra structures). Ontologies are typically constructed in two ways: by extracting information from authoritative textbooks and curriculum standards, or by processing the connections between concepts through natural language, which facilitates input from subject-matter experts [9]. Each node note in the ontology contains metadata, which represent mathematical concepts. These metadata include cognitive complexity, Bloom's taxonomy level and relevance to subsequent topics. The logical progression of math learning is demonstrated in directed acyclic graphs that encode prerequisite relationships to edges.

The ontology facilitates semantic representation and supports multi-relational linking, including contextual and direct dependencies. This facilitates the integration of multimodal student data (including written explanations, assessment results, and problem-solving traces) with conceptual maps. A typical subgraph of the constructed math ontology, shown in Figure 1, illustrates the extensive interconnectivity and nested structure of advanced math courses.

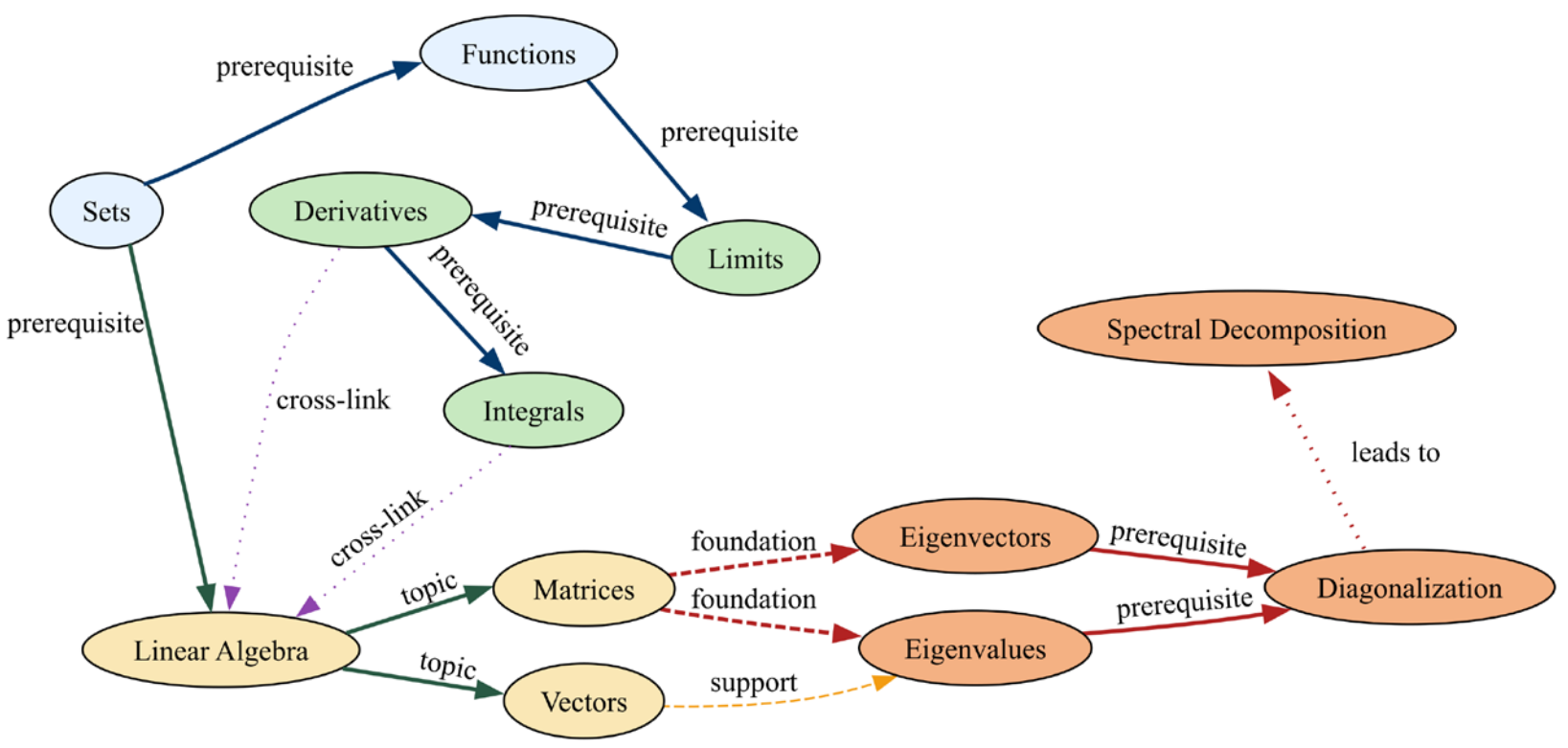

**Figure 1.** Example of Mathematics Concept Ontology

### 2.2 Edge Weighting Based on Problem Difficulty

Quantifying the difficulty of conceptual transitions is essential for adaptive analytics and precision intervention. Each edge in the ontology is weighted to reflect the empirical and theoretical challenge of moving between concepts. The weighting process synthesizes aggregated student performance, problem complexity metadata, and expert-assigned difficulty indices. For any directed edge from concept $i$ to concept $j$, the difficulty coefficient is computed as:

$$D_{ij} = \alpha \cdot \frac{1}{N}\sum_{k=1}^{N}\left(1 - \frac{s_{k,j}}{m_j}\right) + \beta \cdot \delta_{ij} + \gamma \cdot \epsilon_j \tag{1}$$

Here, $s_{k,j}$ is the normalized score of students $k$ on problems for concept $j, m_j$ is the maximum score, $N$ is the student count, $\delta_{ij}$ is an expert difficulty index, and $\epsilon_j$ aggregates complexity features. The coefficients $\alpha, \beta, \gamma$ adjust the influence of each component. Using this approach ensures that edge weights are responsive to instructional priorities and identified learning disabilities [10].

### 2.3 Temporal Graph Updating Mechanism

Students' practice, interactions, and assessments affect their math knowledge. The knowledge graph is updated in real time and the probability of mastery of each node is recalibrated through a Bayesian framework. To ensure that representations remain current, this process integrates new assessment results, assignment submissions, and interaction traces. To model knowledge retention and decay, a temporal decay function is embedded in the update: for concept node $j$, mastery at time $t$ is:

$$M_j(t) = M_j(t-1) \cdot e^{-\lambda \Delta t} + \eta \cdot gain(t) \tag{2}$$

where $M_j(t-1)$ is prior mastery, $\lambda$ is decay rate, $\Delta t$ the elapsed time, and $\eta \cdot$ gain $(t)$ the learning increment from new data. This helps to accurately predict and discover learning gaps [11]. The updating process leads to recalibration of edge weights to reflect changes in the difficulty of discovery. Figure 2 shows the temporal evolution of the student knowledge graph. Node mastery and edge strength vary over the teaching and assessment cycle [12].

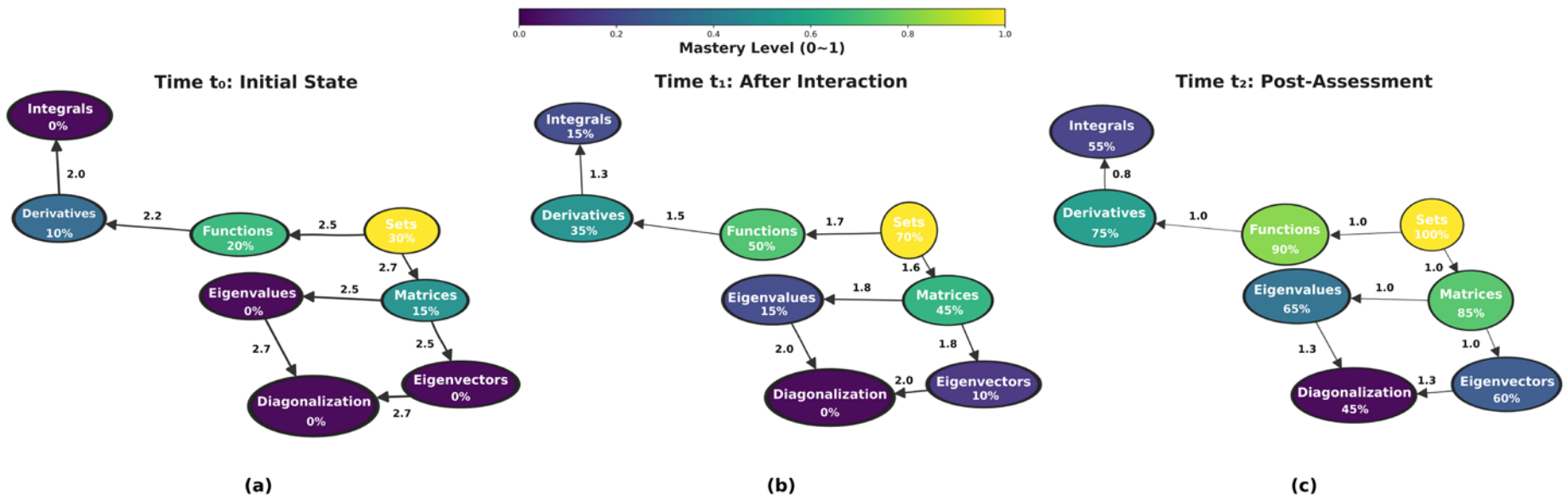

**Figure 2.** Dynamic Knowledge Graph Evolution Process

# 3 Dynamic Prediction Framework

## 3.1 Heterogeneous Graph Attention Layers

Heterogeneous graphical attention mechanisms and temporal modeling are introduced to accurately model the intricate multimodal dependencies in student learning trajectories. The core of this architecture is a stack of graph attention layers, each designed to dynamically recalibrate the influence of neighboring nodes and edges according to both structural and contextual signals. Unlike conventional graph convolution, which treats all neighbors uniformly, the attention mechanism assigns learnable, instance specific weights, enabling nuanced propagation of multimodal feature information.

The attention coefficient for each node-edge-node triplet is computed as follows. Given a node $v_i$ and its neighbor $v_j$, with associated feature vectors $\mathbf{h}_i$ and $\mathbf{h}_j$, and edge feature $\mathbf{e}_{ij}$, the unnormalized attention score is defined by a nonlinear compatibility function:

$$\alpha_{ij} = \frac{exp\left(\phi(W_h \boldsymbol{h}_i, W_h \boldsymbol{h}_j, W_e \boldsymbol{e}_{ij})\right)}{\sum_{k \in \mathcal{N}(i)} exp\left(\phi(W_h \boldsymbol{h}_i, W_h \boldsymbol{h}_k, W_e \boldsymbol{e}_{ik})\right)} \tag{3}$$

where $W_h$ and $W_e$ are learnable weight matrices, and $\phi$ is a composite function integrating bilinear and multi-head projections. To capture higher-order dependencies, we further introduce cross-layer attention regularization, enforcing consistency in the learned attention distribution across the network's depth.

Contextual propagation of node representations over $L$ graph attention layers is performed by recursively aggregating weighted neighbor embeddings:

$$z_i^{(l+1)} = \sigma\left(\sum_{j\in\mathcal{N}(i)} \alpha_{ij}^{(l)} \cdot \left(\boldsymbol{h}_j^{(l)} + \psi(\boldsymbol{e}_{ij})\right)\right) \tag{4}$$

where $\sigma$ is a nonlinear activation and $\psi$ projects the edge features into the node space. Edge update dynamics are similarly governed by attention-weighted feature interactions, allowing the model to adaptively learn the salience of various multimodal cues under different learning scenarios.

The aggregate graph embedding initializes the hidden state of the LSTM module to facilitate temporal embedding transformation. As shown in Figure 3, the output from the stacked graph attention layers is directly connected to the sequence features extracted by the LSTM module. Figure 3 illustrates the complete GAT-LSTM process for robust prediction of learning behavior [13], clearly highlighting the integration of graph-based and sequence-based representations.

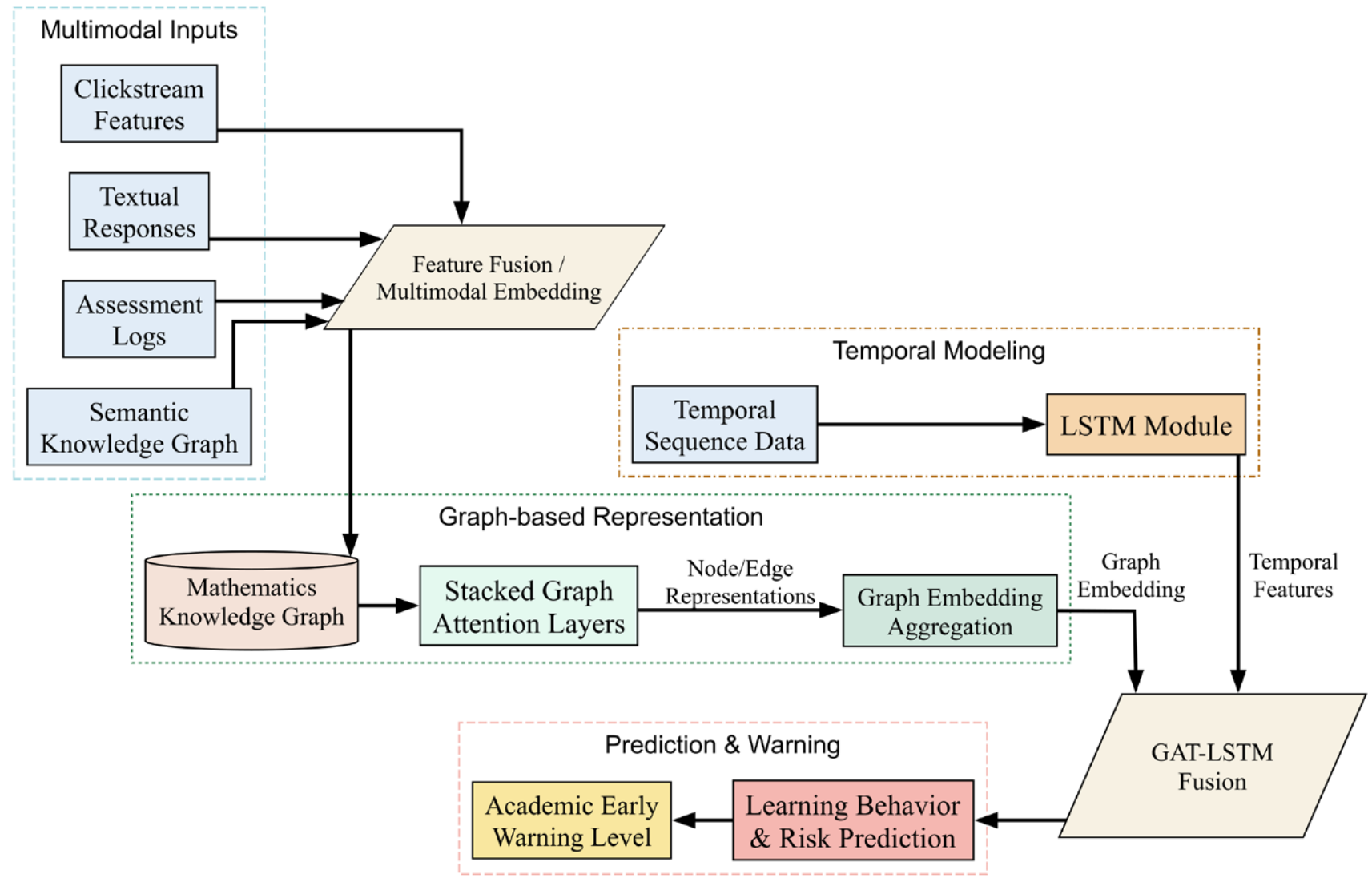

**Figure 3.** GAT-LSTM Architecture Overview

### 3.2 Path-Based Failure Propagation

Modeling the propagation of failure along conceptual paths is crucial for identifying latent vulnerabilities in student knowledge networks. To this end, we formalize a probabilistic error diffusion process on the knowledge graph. Let $\mathcal{P}_{i\to j}$ denote a directed path from concept $i$ to concept $j$, and let $\xi_{ij}(t)$ denote the conditional probability that an error at node $i$ at time $t$ leads to a failure at node $j$ within a bounded time window.

The failure propagation probability along a path of length $n$ is recursively defined as:

$$\xi_{i_1 i_n}(t) = 1 - \prod_{k=1}^{n-1}\left(1 - \omega_{i_k i_{k+1}}(t) \cdot \xi_{i_{k+1} i_n}(t+1)\right) \tag{5}$$

where $\omega_{i_k i_{k+1}}(t)$ is the normalized, time-dependent transmission coefficient capturing the likelihood of failure transfer between consecutive concepts. This recursive relationship is naturally analogous to a cascading effect: an initial error in one base concept leads to the failure of the next subordinate concept.

To account for multimodal uncertainty, we introduce an entropy-regularized diffusion kernel. The expected failure spread $\mathcal{F}_j(t)$ at node $j$ is given by:

$$\mathcal{F}_j(t) = \sum_{i \in \mathcal{A}(j)} \pi_i(t) \cdot \xi_{ij}(t) + \lambda H\left(\{\xi_{ij}(t)\}\right) \tag{6}$$

where $\mathcal{A}(j)$ is the set of all ancestor nodes of $j, \pi_i(t)$ is the empirical error rate at node $i$, and $H$ is the Shannon entropy. The parameter $\lambda$ balances deterministic pathwise propagation and stochastic uncertainty. This framework allows for the identification of critical failure chains and quantifies the risk associated with particular learning trajectories [14].

### 3.3 Adaptive Warning Level Classification

To translate predictive outputs into actionable academic early warnings, we propose a dynamic thresholding mechanism for warning level classification, which adapts in real time to the evolving distribution of risk scores. Each student's risk vector $\mathbf{r}_t$, derived from the fused graphsequence model, is mapped to a set of tiered warning levels using a soft boundary optimization approach.

Formally, the optimal set of thresholds $\{\tau_k\}$ is determined by minimizing an empirical risk function:

$$\min_{\{\tau_k\}} \sum_{i=1}^{N} \sum_{k=1}^{K} \mathbb{I}\left(\tau_{k-1} < r_{i,t} \leq \tau_k\right) \cdot \mathcal{L}\left(y_{i,t}, w_k\right) + \beta \sum_{k=1}^{K-1} |\tau_k - \tau_{k-1} - \mu_k|^2 \tag{7}$$

where $\mathbb{I}$ is the indicator function, $y_{i,t}$ the ground-truth warning level, $w_k$ the class label for threshold $k$, and $\mu_k$ the target interval between warning levels. The regularization term, weighted by $\beta$, promotes well-separated, stable thresholds.

In addition, thresholds are periodically recalibrated through a moving window Bayesian updating method to ensure that warnings remain sensitive and accurate as population risk patterns evolve. This helps to minimize the drift of thresholds in the presence of class imbalances and shifts in the concept of time. This adaptive mechanism facilitates rapid, individualized academic interventions and reduces false alarms for at-risk students [15].

## 4 Case Study

### 4.1 Eigenvalue Problem Patterns

A semester-long survey was conducted among undergraduate mathematics students, focusing on their understanding of eigenvalues and eigenvectors, to validate the dynamic prediction framework. A multimodal dataset consisting of score assessments, step-by-step problem-solving logs, and temporal engagement data was utilized to model knowledge trajectories at high resolution. The knowledge graph maps the probability of student mastery to nodes associated with eigenvalues. The focus of the mapping is to identify potential dependencies that may lead to error propagation.

A high-dimensional heatmap was created to demonstrate mastery for each eigenvector problem set. Each cell in the heat map represents a student-concept pair, and the color intensity represents the posterior probability of mastery computed by the Bayesian updating mechanism. The heat map revealed a number of important findings. First, the failure of matrix multiplication and diagonalization prerequisites is consistent with significant clustering of low mastery cases. This suggests that early conceptual differences propagate and impede eigenvector understanding. Second, attention-weighted edge analysis revealed a dense cascade of errors for some students in the spectral decomposition subgraph. This implies that targeted remediation for this concept may yield considerable learning gains.

Figure 4 shows a heat map of mastery for a representative group of students, indicating the high-risk group. On the other hand, the vertical bands of lower mastery emphasize the interrelatedness of conceptual dependencies. The dynamic updating of the heatmap over successive weeks further revealed the positive effect of timely interventions, as regions of low mastery receded following targeted instructional support.

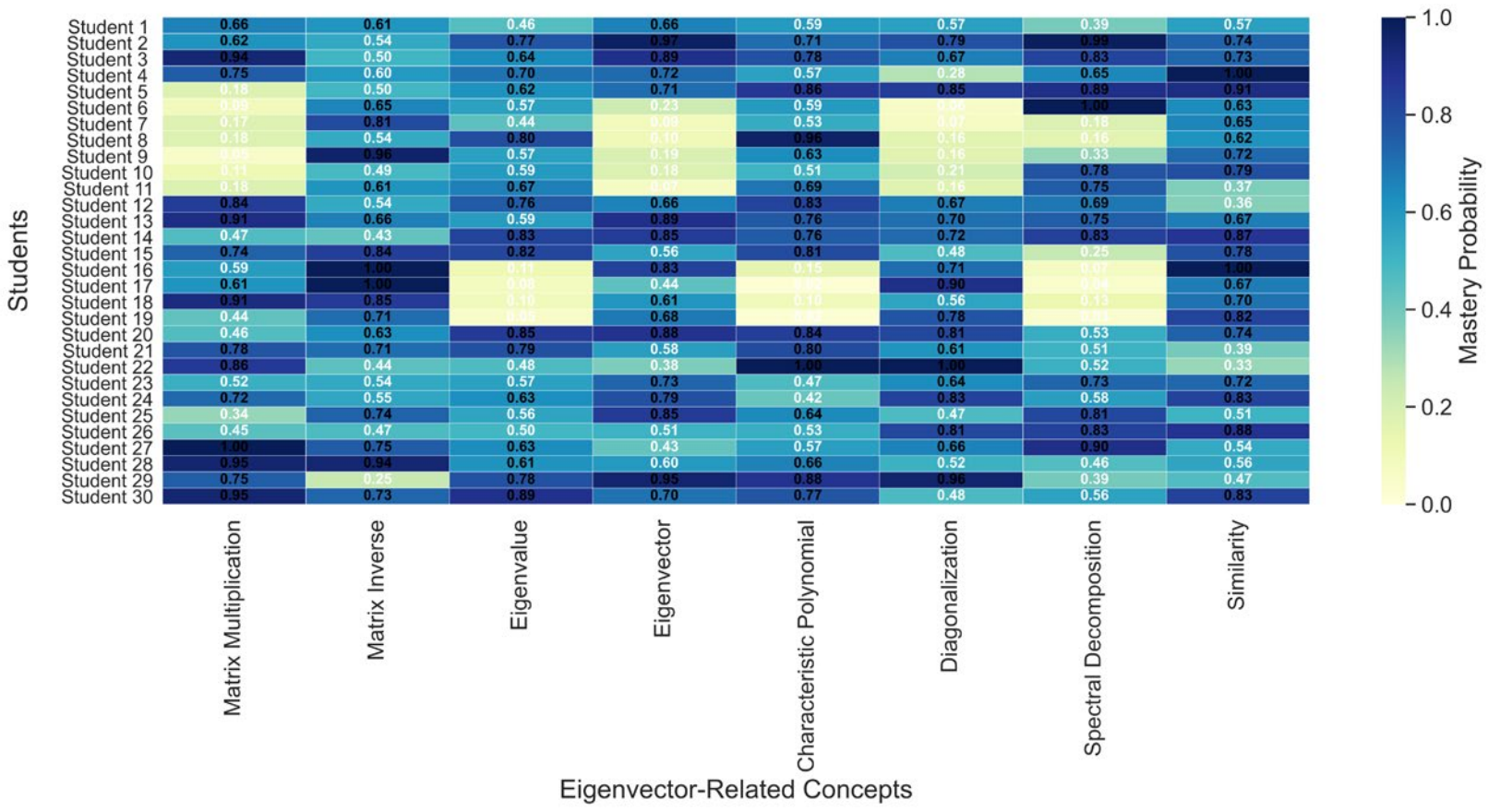

**Figure 4.** Student Graph Visualization

To enhance interpretability and provide actionable feedback, we visualized the individual knowledge graphs for selected students before and after adaptive intervention. In these graphs, each node represents a mathematical concept, and the thickness of each edge indicates the current transition difficulty, dynamically weighted according to student data.

Before the intervention, the knowledge graphs typically exhibited fragmented mastery, weak subgraph connectivity, and numerous high-difficulty edges linking key eigenvalue concepts to their prerequisites. These characteristics suggest that students struggled with foundational concepts, leading to isolated knowledge clusters and significant learning barriers. After the intervention, however, the graphs displayed a denser backbone, fewer isolated nodes, and stronger inter-concept connectivity. This visual transition reflects the consolidation of foundational knowledge and the successful bridging of key conceptual gaps. As illustrated in Figure 5, these changes are clearly visible for two representative students. The pre-intervention state is marked by sparse and disconnected graph structures, while the post-intervention state demonstrates a robust, well-connected network of mathematical concepts.

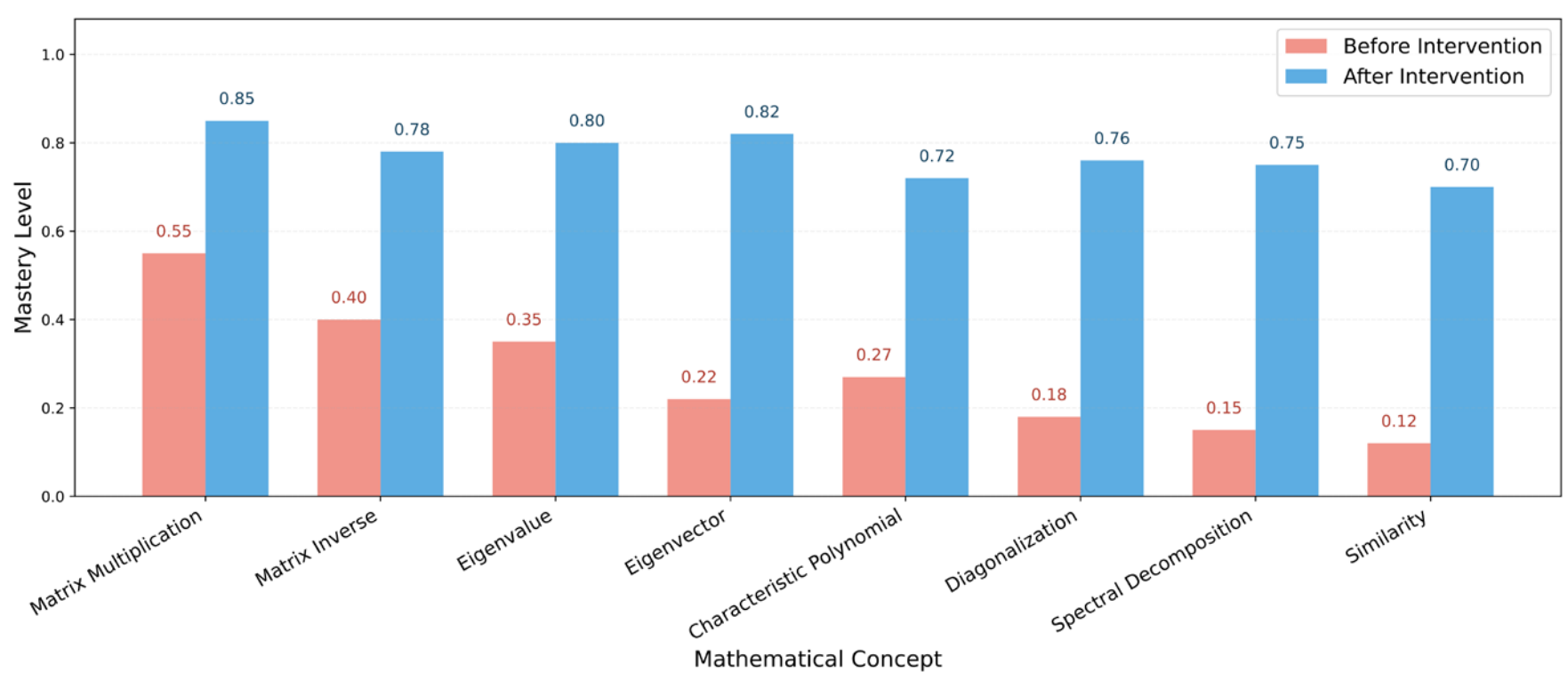

**Figure 5.** Student Knowledge Graphs Before (left) and After (right) Intervention.

To rigorously assess the impact of the intervention, a sample group of 10 high-risk students was further analyzed for their knowledge mastery and conceptual connectivity. The results show that the average mastery probability across key mathematical concepts increased from 0.43 (pre-intervention) to 0.71 (post-intervention), representing an improvement of approximately 65%. Meanwhile, the average edge weight-representing the difficulty of transitions between core concepts-decreased from 0.38 to 0.21, indicating that conceptual barriers were significantly reduced after intervention. Notably, within the eigenvalue decomposition subgraph, the mastery probability for the "matrix diagonalization" node increased from 0.35 to 0.68.

The percentage improvements shown here (e.g., $+65\%$ for average mastery probability and $+94.3\%$ for "matrix diagonalization" mastery) are calculated as relative increases from pre- to post-intervention, i.e., (post - pre)/pre $\times$ $100\%$. For example, the average mastery probability increased from 0.43 to 0.71, which corresponds to a $65\%$ improvement $[(0.71 - 0.43)/0.43 \times 100\% \approx 65\%]$, and the mastery rate for "matrix diagonalization" rose from 0.35 to 0.68, yielding a 94.3% increase.

These improvements are not only apparent in the visualization but are also supported by quantitative metrics, as summarized in Table 1 below:

**Table 1.** Quantitative comparison of student knowledge graph metrics before and after intervention.

| Metric | Pre-intervention | Post-intervention | Improvement |
| --- | --- | --- | --- |
| Average mastery probability | 0.43 | 0.71 | +65% |
| Average key edge weight | 0.38 | 0.21 | -44.7% |
| "Matrix diagonalization" mastery | 0.35 | 0.68 | +94.3% |

These findings suggest that the proposed framework not only enables the dynamic visualization of student knowledge states but also provides quantitative evidence for the effectiveness of targeted instructional interventions. Teachers can utilize these visual and data-driven insights to identify specific conceptual barriers, monitor students' progress, and design personalized learning strategies that effectively improve learning outcomes.

### 4.2 Intervention Strategies

The effectiveness of the intervention was assessed using the instructional strategy utility function in the context of expected learning gains and risk mitigation under different adaptive policies. Let $U_{\text{gain}}(S,t)$ denote the cumulative knowledge gain for student $S$ at time $t$, defined as

$$U_{gain}(S,t) = \sum_{j \in \mathcal{C}} w_j \left[ M_j^{post}(S,t) - M_j^{pre}(S,t) \right] \tag{8}$$

where $\mathcal{C}$ is the set of targeted concepts, $w_j$ is the domain-weighted importance of concept $j$, and $M_j^{pre}, M_j^{post}$ are mastery levels before and after intervention.

To assess risk reduction, we define the risk utility function

$$U_{risk}(S,t) = -\sum_{j \in \mathcal{C}} \gamma_j \left[ \mathcal{F}_j^{post}(S,t) - \mathcal{F}_j^{pre}(S,t) \right] \tag{9}$$

where $\gamma_j$ scales the penalty of failure propagation, and $\mathcal{F}_j^{pre}, \mathcal{F}_j^{post}$ are the expected failure probabilities on concept $j$ before and after intervention.

The composite intervention benefit is then captured by the optimization objective

$$U_{total}(S,t) = \lambda_1 U_{gain}(S,t) + \lambda_2 U_{risk}(S,t) - \lambda_3 C_{int}(S,t) \tag{10}$$

where $C_{\text{int}}$ denotes the instructional cost, and $\lambda_1, \lambda_2, \lambda_3$ are tunable weights expressing institutional priorities. This formalism lays the foundation for selecting, deploying, and evaluating adaptive instructional strategies. These strategies ensure that the allocation of resources in complex, high-stakes math learning environments both reduces risk and improves student achievement.

To provide a more comprehensive understanding of the intervention strategy outcomes, the study tracked and analyzed both the implementation process and specific student cases. The intervention process began with the identification of at-risk students through the early warning outputs of the knowledge graph model, which pinpointed students with low mastery probabilities and high risk of error propagation. Once identified, these students received tailored support, including focused one-on-one tutoring and adaptive assignments dynamically generated according to their individual knowledge gaps. Throughout the intervention, students were provided with visual feedback through their personal knowledge graphs, which helped them recognize progress and target remaining weaknesses.

For example, one high-risk student who consistently struggled with matrix diagonalization and related foundational concepts participated in several rounds of targeted tutoring and adaptive practice. As a result, the student's mastery probability for matrix diagonalization increased notably from 0.32 to 0.69, and the average transition difficulty between core concepts was reduced by more than half. Across the intervention group, the average mastery probability rose from 0.43 to 0.71, while transition difficulties between key concepts decreased from 0.38 to 0.21. Notably, the mastery rate for matrix diagonalization improved by 94.3%. The risk utility function also reflected an average reduction of 0.19 in the probability of failure propagation along critical conceptual paths.

Follow-up analysis four weeks post-intervention demonstrated that most students maintained their improved mastery levels and were less likely to return to at-risk status. Both quantitative outcomes and qualitative feedback from students and instructors confirmed that the intervention strategies not only closed immediate learning gaps, but also fostered increased engagement, confidence, and long-term academic resilience.

# 5 Conclusion

This study proposed an innovative, dynamic framework for predicting learning behaviors and issuing academic early warnings in advanced mathematics by integrating multimodal data analysis and knowledge graph techniques. The empirical results demonstrate that this approach can reliably identify students at risk, accurately trace the spread of conceptual errors, and provide actionable guidance for timely and effective intervention.

The implementation of targeted intervention strategies yielded significant and sustained improvements in student outcomes. After receiving individualized tutoring, adaptive assignments, and graph-driven feedback, high-risk students achieved substantial gains in conceptual mastery and academic stability. For example, the average mastery probability across key mathematical concepts increased from 0.43 to 0.71, while the difficulty of conceptual transitions, as reflected in edge weights, decreased notably. Mastery of matrix diagonalization—a key area of difficulty—nearly doubled after intervention. Furthermore, the overall risk of error propagation was reduced, as indicated by a drop in the risk utility score.

A representative case highlighted that a student who had previously struggled with matrix diagonalization benefited greatly from focused, adaptive support, exhibiting a significant increase in mastery and a reduction in conceptual barriers. Longitudinal tracking over a month after intervention revealed that the majority of students sustained their progress, with few returning to "at-risk" status. Feedback from both students and instructors underscored the value of the intervention framework in enhancing understanding, motivation, and confidence.

In summary, the integration of multimodal analytics, temporal modeling, and personalized intervention strategies not only led to immediate improvements in learning outcomes but also supported long-term academic development and resilience in ad-vanced mathematics. These findings support the adoption of data-driven, adaptive intervention frameworks to promote more effective and individualized mathematics education.

## 6 FUNDING

This work was supported by the second batch of provincial-level "Golden Course" offline first-class courses in Guizhou Province: Advanced Mathematics (number: 2024JKXX0218), and the 2024 Guizhou Province Undergraduate Teaching Content and Curriculum System Reform Project: Research on the Reform of the "BOPPPS+Learning Pass" Blended Teaching Mode Based on the Student-centered Concept (number: GZJG2024445).